\documentclass{article}
\pdfoutput=1

\usepackage{my_style}

\usepackage[utf8]{inputenc} 
\usepackage[T1]{fontenc}    
\usepackage{times}

\usepackage[round]{natbib}
\usepackage{graphicx}
\usepackage{rotating}
\usepackage{array}
\usepackage{algorithm}
\usepackage{algorithmic}
\usepackage[labelfont=bf]{caption}
\usepackage{subcaption}

\usepackage{amsmath,amsfonts,amssymb,bbm,epsfig,bm}

\usepackage{theorem}
\newcommand{\BlackBox}{\rule{1.5ex}{1.5ex}}  

\newcommand{\cbr}[1]{\left\{#1\right\}}

\renewcommand{\algorithmicrequire}{\textbf{Input:}}
\renewcommand{\algorithmicensure}{\textbf{Output:}}

\newcommand{\E}{\mathbb{E}}
\newcommand{\KLD}{\text{KL}}
\newcommand{\ENT}{\text{H}}
\newcommand{\comm}{\text{comm}}
\newcommand{\indicator}{\mathbbm{1}}

\newcolumntype{L}[1]{>{\raggedright\arraybackslash}p{#1}}
\newcolumntype{R}[1]{>{\raggedleft\arraybackslash}p{#1}}
\newcolumntype{C}[1]{>{\centering\let\newline\\\arraybackslash\hspace{0pt}}m{#1}}
\newcolumntype{?}{!{\vrule width 1pt}}
\makeatletter
\newcommand{\thickhline}{%
    \noalign {\ifnum 0=`}\fi \hrule height 1pt
    \futurelet \reserved@a \@xhline
}
\newcolumntype{"}{@{\hskip\tabcolsep\vrule width 1pt\hskip\tabcolsep}}
\makeatother

\DeclareMathOperator*{\argmax}{\mathrm{argmax}}

\newcommand{\intset}[1]{\cbr{1..n}}

\usepackage[colorlinks,pdfpagelabels,plainpages=false]{hyperref}
\usepackage{rotating}

\usepackage{xcolor}
\definecolor{dark-red}{rgb}{0.4,0.15,0.15}
\definecolor{dark-blue}{rgb}{0.15,0.15,0.4}
\definecolor{medium-blue}{rgb}{0,0,0.5}
\hypersetup{
   colorlinks, linkcolor={dark-blue},
   citecolor={dark-blue}, urlcolor={medium-blue}
}

\usepackage{booktabs}
\usepackage{bm}
\usepackage{amsmath}
\usepackage{amssymb}
\usepackage{amsfonts}
\usepackage{mathtools}
\usepackage{comment}

\newcommand{\mbf}[1]{{\boldsymbol{\mathbf{#1}}}}
\renewcommand{\bm}{\mbf}

\usepackage{color}

\usepackage{parskip}

\usepackage{multirow}
\usepackage{titlefoot}
\usepackage{gensymb}

\title{\bf Connecting the Dots Between MLE and RL for Sequence Prediction}

\renewcommand{\footnotesize}{\normalsize}

\author{
{\normalsize 
Bowen Tan$^{1*}$,~~ Zhiting Hu$^{1,2*}$,~~ Zichao Yang$^{1}$,~~ Ruslan Salakhutdinov$^{1}$,~~ Eric P. Xing$^{1,2}$
}
\and
{\normalsize
(*equal contribution)~~~
Carnegie Mellon University$^1$,~ Petuum Inc.$^2$ 
}\\
{\normalsize
\texttt{\{zhitinghu,bwkevintan,yangtze2301\}@gmail.com}
}
\\
{\small
\texttt{rsalakhu@cs.cmu.edu}\quad \texttt{eric.xing@petuum.com}
}
}

\begin{document}
\date{}

\maketitle

\begin{abstract}

\begin{sloppypar}
Sequence prediction models can be learned from example sequences with a variety of training algorithms. Maximum likelihood learning is simple and efficient, yet can suffer from compounding error at test time. 
Reinforcement learning such as policy gradient addresses the issue but can have prohibitively poor exploration efficiency. A rich set of other algorithms such as RAML, SPG, and data noising, have also been developed from different perspectives. This paper establishes a formal connection between these algorithms. We present a generalized entropy regularized policy optimization formulation, and show that the apparently distinct algorithms can all be reformulated as special instances of the framework, with the only difference being the configurations of a reward function and a couple of hyperparameters. The unified interpretation offers a systematic view of the varying properties of exploration and learning efficiency. Besides, inspired from the framework, we present a new algorithm that dynamically interpolates among the family of algorithms for scheduled sequence model learning. Experiments on machine translation, text summarization, and game imitation learning demonstrate the superiority of the proposed algorithm.
\end{sloppypar}

\end{abstract}

\section{Introduction}\label{sec:intro}
Sequence prediction problem is ubiquitous in many applications, such as generating a sequence of words for machine translation~\citep{wu2016google,sutskever2014sequence}, text summarization~\citep{hovy1998automated,rush2015neural}, and image captioning~\citep{vinyals2015show,karpathy2015deep}, or taking a sequence of actions to complete a task.
In these problems~\citep[e.g.,][]{mnih2015human,ho2016generative}, we are often given a set of sequence examples, from which we want to learn a model that sequentially makes the next prediction (e.g., generating the next token) given the current state (e.g., the previous tokens). 

A standard training algorithm is based on supervised learning which seeks to maximize the log-likelihood of example sequences (i.e., maximum likelihood estimation, MLE). 
Despite the computational simplicity and efficiency, MLE training can suffer from compounding error~\citep{ranzato2015sequence,ross2010efficient} in that mistakes at test time accumulate along the way and lead to states far from the training data. Another line of approaches overcome the training/test discrepancy issue by resorting to the reinforcement learning (RL) techniques~\citep{ranzato2015sequence,bahdanau2016actor,rennie2017self}. For example, \citet{ranzato2015sequence} used policy gradient~\citep{sutton2000policy} to train a text generation model with the task metric (e.g., BLEU) as reward.
However, RL-based approaches can face challenges of prohibitively poor sample efficiency and high variance. To this end, a diverse set of methods has been developed that is in a middle ground between the two paradigms of MLE and RL. For example, RAML~\citep{norouzi2016reward} adds reward-aware perturbation to the MLE data examples; SPG~\citep{ding2017cold} leverages reward distribution for effective sampling of policy gradient. Other approaches such as data noising~\citep{xie2017data} also show improved results.

\begin{figure}[t]
\centering
\includegraphics[width=0.8\textwidth]{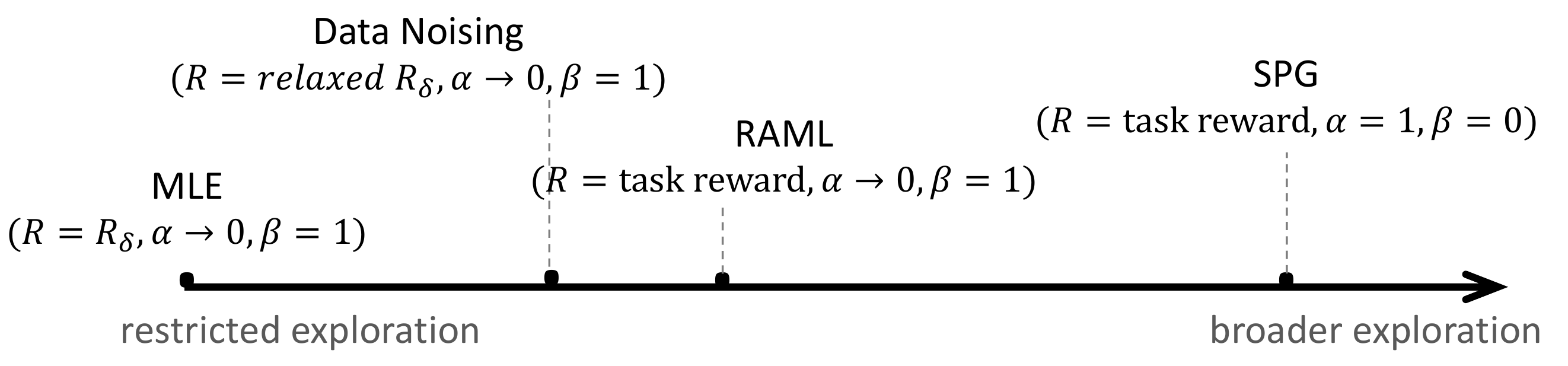}
\caption{Unified formulation of different learning algorithms. Each algorithm is a special instance of the general ERPO framework by taking different specification of the hyperparameters $(R, \alpha, \beta)$ in Eq.\eqref{eq:pg}.}
\label{fig:conn-dots}
\vspace{-12pt}
\end{figure}

In this paper, we establish a unified perspective of the above distinct learning algorithms. Specifically, we present a generalized entropy regularized policy optimization framework, and show that the diverse algorithms, including MLE, RAML, SPG, and data noising, can all be re-formulated as special cases of the framework, with the only difference being the choice of reward and the values of a couple of hyperparameters (Figure~\ref{fig:conn-dots}). In particular, we show MLE is equivalent to using a \emph{Delta}-function reward which returns 1 to model samples that match training examples exactly, and $-\infty$ to any other samples. Such extremely restricted reward has literally disabled any exploration of the model beyond training data, yielding brittle prediction behaviors. Other algorithms essentially use various locally-relaxed rewards, joint with the model distribution, for broader (and more costly) exploration during training. 

Besides the new views of the existing algorithms, the unified perspective also inspires new algorithms for improved learning. We present an interpolation algorithm as a direct application of the framework. As training proceeds, the algorithm gradually expands the exploration space by annealing the reward and hyperparameter values. The annealing in effect dynamically interpolates among the existing algorithms from left to right in Figure~\ref{fig:conn-dots}. We conduct experiments on the tasks of text generation including machine translation and text summarization, and game imitation learning. The interpolation algorithm shows superior performance over various previous methods.


%


%


\section{Related Work}\label{sec:related}
Given a set of data examples, sequence prediction models are usually trained to maximize the log-likelihood of the next label (token, action) conditioning on the current state observed in the data. 
Reinforcement learning (RL) addresses the discrepancy between training and test by also using models' own predictions at training time. Various RL approaches have been applied for sequence generation, such as policy gradient~\citep{ranzato2015sequence} and actor-critic~\citep{bahdanau2016actor}. Softmax policy gradient (SPG)~\citep{ding2017cold} additionally incorporates the reward distribution to generate high-quality sequence samples. The algorithm is derived by applying a log-softmax trick to adapt the standard policy gradient objective. 
Reward augmented maximum likelihood (RAML)~\citep{norouzi2016reward} is an algorithm in between MLE and policy gradient. It is originally developed to go beyond the maximum likelihood criteria and incorporate task metric (such as BLEU for machine translation) to guide the model learning. Mathematically, RAML shows that MLE and maximum-entropy policy gradient are respectively minimizing a KL divergences in opposite directions. \citet{koyamada2018alpha} thus propose to use the more general $\alpha$-divergence as a combination of the two paradigms. Our framework is developed in a different perspective, reformulates a different and more comprehensive set of algorithms, and leads to new insights in terms of exploration and learning efficiency of the various algorithms.
Besides the algorithms discussed in the paper, there are other learning methods for sequence models. For example, \citet{hal2009search,leblond2017searnn,wiseman2016sequence} use a learning-to-search paradigm for sequence  generation or structured prediction. Scheduled Sampling~\citep{bengio2015scheduled} adapts MLE by randomly replacing ground-truth tokens with model predictions as the input for decoding the next-step token. 
\citet{hu2017controllable,yang2018unsupervised,fedus2018maskgan} learn (conditional) text generation with holistic, structured discriminators. \citet{zhu2018text} explore the new setting of text infilling that leverages both left- and right-side context for generation.

Policy optimization for reinforcement learning is studied extensively in robotic and game environment. For example, \citet{peters2010relative} introduce a relative entropy regularization to reduce information loss during learning. \citet{schulman2015trust} develop a trust-region approach for monotonic improvement. \citet{dayan1997using,levine2018reinforcement,abdolmaleki2018maximum} study the policy optimization algorithms in a probabilistic inference perspective. 
\citet{hu2018deep} show the connections between policy optimization, Bayesian posterior regularization~\citep{hu2016harnessing,ganchev2010posterior}, and GANs~\citep{goodfellow2014generative} for combining structured knowledge with deep generative models.
The entropy-regularized policy optimization formulation presented here can be seen as a generalization of many of the previous policy optimization methods. Besides, we formulate the framework primarily in the sequence generation context.





\section{Connecting the Dots}\label{sec:framework}
We first present a generalized formulation of an entropy regularized policy optimization framework. Varying the reward and hyperparameter values of the framework instantiates different existing algorithms that were originally developed in distinct perspectives. We discuss the effect of the reward and hyperparameter values on exploration and computation efficiency for model learning, and thus provide a consistent view of the family of algorithms. 

For clarity, we present the framework in the sequence generation context. The formulations can straightforwardly be extended to other settings such as imitation learning in robotic and game environments, as discussed briefly at the end of this section and shown in the experiment.

We first establish the notations. Let $\bm{y}=(y_1, \dots, y_T)$ be the sequence of $T$ tokens. Let $\bm{y}^*$ be a training example drawn from the empirical data distribution.
From the sequence examples, we aim to learn a sequence generation model $p_\theta(\bm{y})=\prod_t p_\theta(y_t|\bm{y}_{1:t-1})$ with parameters $\bm{\theta}$. Note that generation of $\bm{y}$ can condition on other factors. For example, in machine translation, $\bm{y}$ is the sentence in target language and depends on an input sentence in source language. For simplicity of notations, we omit the conditioning factors.


\subsection{Entropy Regularized Policy Optimization (ERPO)}
Policy optimization is a family of reinforcement learning (RL) algorithms.
Given a reward function $R(\bm{y}| \bm{y}^*)\in\mathbb{R}$  (e.g., BLEU score in machine translation) that evaluates the quality of generation $\bm{y}$ against the true $\bm{y}^*$, the general goal of policy optimization is to learn the model $p_\theta(\bm{y})$ (a.k.a policy) to maximize the expected reward. Previous research on \emph{entropy regularized} policy optimization (ERPO) stabilizes the learning by augmenting the objective with information theoretic regularizers. Here we present a generalized variational formulation of ERPO. More concretely, we assume a non-parametric variational distribution $q(\bm{y})$ w.r.t the model $p_\theta(\bm{y})$. The objective to maximize is then:
\begin{equation}
\small
\begin{split}
\mathcal{L}(q, \bm{\theta}) = \E_{q}\left[ R(\bm{y} | \bm{y}^*) \right] - \alpha \KLD\big( q(\bm{y}) \| p_\theta(\bm{y}) \big) + \beta \ENT(q),
\end{split}
\label{eq:pg}
\end{equation}
where $\KLD(\cdot\|\cdot)$ is the Kullback–Leibler divergence forcing $q$ to stay close to $p_\theta$; $\ENT(\cdot)$ is the Shannon entropy imposing maximum entropy assumption on $q$; and $\alpha$ and $\beta$ are balancing weights of the respective terms. Intuitively, the objective is to maximize the expected reward under the variational distribution $q$ while minimizing the distance between $q$ and $p_\theta$, with a maximum entropy regularizer on $q$. The above formulation is relevant to and can be seen as a variant of previous policy optimization approaches in RL literature, such as relative entropy policy search~\citep{peters2010relative}, maximum entropy policy gradient~\citep{ziebart2010modeling,haarnoja2017reinforcement}, and others where $q$ is formulated either as a non-parametric distribution as ours~\citep{abdolmaleki2018maximum,peters2010relative} or as a parametric one~\citep{schulman2015trust,schulman2017equivalence,teh2017distral}.

The objective can be maximized with an EM-style procedure that iterates two coordinate ascent steps optimizing $q$ and $\bm{\theta}$, respectively. At iteration $n$:
\begin{equation}
\small
\begin{split}
\text{E-step:}\quad &q^{n+1}(\bm{y}) \propto \exp\left\{ \frac{\alpha \log p_{\theta^{n}}(\bm{y}) + R(\bm{y}|\bm{y}^*)}{\alpha + \beta} \right\},\\
\text{M-step:}\quad &\bm{\theta}^{n+1} = \argmax\nolimits_\theta \E_{q^{n+1}} \big[ \log p_\theta(\bm{y}) \big].
\end{split}
\label{eq:pg-em}
\end{equation}
In the E-step, $q$ has a closed-form solution. We can have an intuitive interpretation of its form. First, it is clear to see that if $\alpha\to\infty$, we have $q^{n+1}=p_\theta^n$. This is also reflected in the objective Eq.(\ref{eq:pg}) where the weight $\alpha$ encourages $q$ to be close to $p_\theta$.  Second, the weight $\beta$ serves as the temperature of the $q$ softmax distribution. In particular, a large temperature $\beta\to\infty$ makes $q$ a uniform distribution, which is consistent with the outcome of an infinitely large maximum entropy regularization in Eq.(\ref{eq:pg}).
In the M-step, the update rule can be interpreted as maximizing the log-likelihood of samples from the distribution $q$.

In the context of sequence generation, it is sometimes more convenient to express the equations at token level, as shown shortly. To this end, we decompose $R(\bm{y}|\bm{y}^*)$ along the time steps:
\begin{equation}
\small
\begin{split}
R(\bm{y}|\bm{y}^*) 
= \sum\nolimits_t R(\bm{y}_{1:t}|\bm{y}^*) - R(\bm{y}_{1:t-1}|\bm{y}^*)
:= \sum\nolimits_t \Delta R(y_t | \bm{y}_{1:t-1}, \bm{y}^*),
\end{split}
\label{eq:token-level-reward}
\end{equation}
where $\Delta R(y_t | \bm{y}^*, \bm{y}_{1:t-1})$ measures the reward contributed by token $y_t$. The solution of $q$ in Eq.(\ref{eq:pg-em}) can then be re-written as:
\begin{equation}
\small
\begin{split}
q^{n+1}(\bm{y}) \propto \prod\nolimits_t \exp\left\{\frac{\alpha \log p_{\theta^{n}}(y_{t}|\bm{y}_{1:t-1}) + \Delta R(y_t|\bm{y}_{1:t-1}, \bm{y}^*)}{\alpha + \beta} \right\}
\end{split}
\label{eq:token-level-q}
\end{equation}

The above ERPO framework has three key hyperparameters, namely $(R, \alpha, \beta)$. In the following, we show that different values of the three hyperparameters correspond to different learning algorithms (Figure~\ref{fig:conn-dots}).
In particular, we first connect MLE to the above general formulation, and compare MLE and regular ERPO from the new perspective.

\subsection{MLE as a Special Case of ERPO}\label{sec:mle}
Maximum likelihood estimation is the most widely-used approach to learn a sequence generation model due to its simplicity and efficiency. It aims to find the optimal parameter value that maximizes the data log-likelihood:
\begin{equation}
\small
\begin{split}
\bm{\theta}^* 
= \argmax\nolimits_\theta \mathcal{L}_{\text{MLE}}(\bm{\theta})
= \argmax\nolimits_\theta \log p_\theta(\bm{y}^*).
\end{split}
\label{eq:mle}
\end{equation}

We show that the MLE objective can be recovered from Eq.(\ref{eq:pg-em}) with specific reward and hyperparameter configurations.
Consider a $\delta$-reward defined as\footnote{For token-level, define $R_\delta(\bm{y}_{1:t}|\bm{y}^*) = t/T^*$ if $\bm{y}_{1:t}=\bm{y}^*_{1:t}$ and $-\infty$ otherwise, where $T^*$ is the length of $\bm{y}^*$. Note that the $R_\delta$ value of $\bm{y}=\bm{y}^*$ can also be set to any constant larger than $-\infty$.}:
\begin{equation}
\small
\begin{split}
R_\delta(\bm{y}|\bm{y}^*) = \left\{
  \begin{array}{ll}
  1 & \text{if}\ \ \bm{y} = \bm{y}^* \\
  -\infty & \text{otherwise}.
  \end{array}
  \right.
\end{split}
\label{eq:delta-reward}
\end{equation}
Let $(R=R_\delta, \alpha\to 0, \beta=1)$ in the framework. From the E-step of Eq.(\ref{eq:pg-em}), we have $q(\bm{y}|\bm{x}) = 1$ if $\bm{y}=\bm{y}^*$ and $0$ otherwise. The M-step is therefore equivalent to $\argmax_\theta \log p_\theta(\bm{y}^*|\bm{x})$, which recovers precisely the MLE objective in Eq.(\ref{eq:mle}). (Note that the very small $\alpha$ is still $>0$, making the M-step for maximizing the objective Eq.\eqref{eq:pg} valid and necessary.)

That is, MLE can be seen as an instance of the policy optimization with the $\delta$-reward and the specialized $(\alpha, \beta)$ values. Any sample $\bm{y}$ that fails to match the given data $\bm{y}^*$ exactly will receive a negative infinite reward and never contribute to model learning. 

\paragraph{Exploration efficiency}\quad

\begin{figure*}[t]
    \centering
    \begin{subfigure}[b]{0.32\textwidth}
        \centering
        \includegraphics[width=\textwidth]{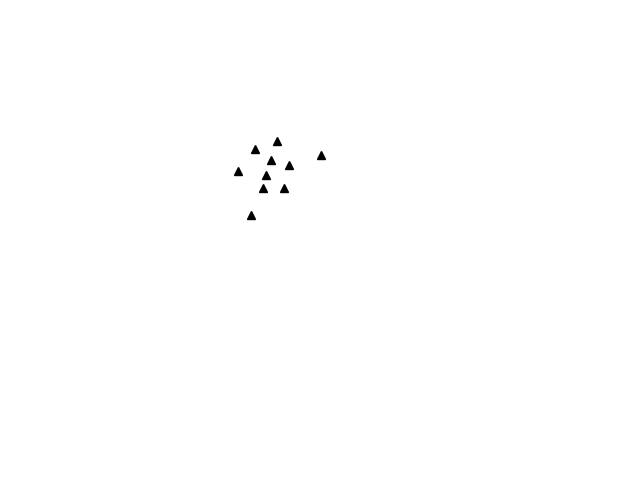}
        \vspace{-25pt}
        \caption{}
    \end{subfigure}
    \begin{subfigure}[b]{0.32\textwidth}
        \centering
        \includegraphics[width=\textwidth]{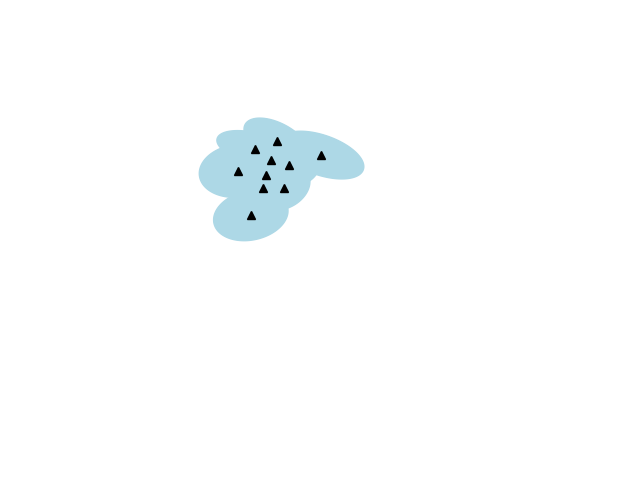}
        \vspace{-25pt}
        \caption{}
    \end{subfigure}%
    \begin{subfigure}[b]{0.32\textwidth}
        \centering
        \includegraphics[width=\textwidth]{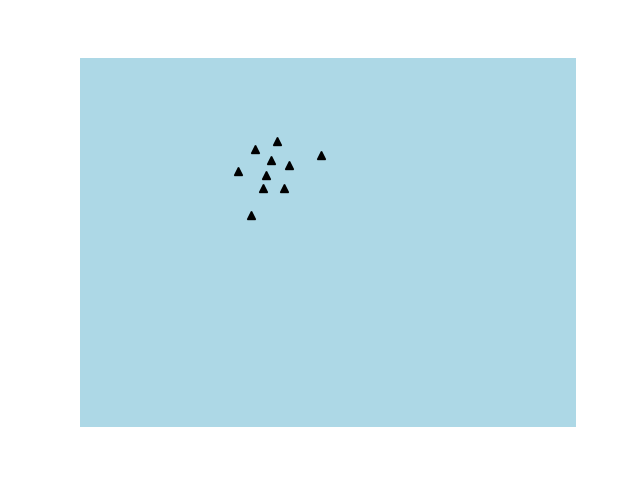}
        \vspace{-25pt}
        \caption{}
    \end{subfigure}%
    \caption{Exploration space exposed for model learning in different algorithms. {\bf (a)}: The valid exploration space of MLE is exactly the set of training examples. {\bf (b)}: RAML and Data Noising use diffused rewards and allow larger exploration space surrounding the training examples. {\bf (c)}: Common policy optimization such as SPG basically allows the whole exploration space.}
\label{fig:space}
\end{figure*}

The ERPO reformulation of MLE enables us to view the characteristics of the algorithm in term of exploration efficiency. 
Concretely, the $\delta$-reward has permitted only samples that match training examples, and made invalid any exploration beyond the small set of training data (Figure~\ref{fig:space}(a)). The extremely restricted exploration at training time results in a brittle model that can easily encounter unseen states and make mistakes in prediction.

On the other hand, however, a major advantage of the $\delta$-reward is that it defines a distribution over the sequence space from which sampling is reduced to simply picking an instance from the training set. The resulting samples are ensured to have high quality.
This makes the MLE implementation very simple and the computation efficient in practice.

On the contrary, common rewards (e.g., BLEU) used in policy optimization are more diffused than the $\delta$-reward, and thus allow exploration in a broader space with valid reward signals. However, the diffused rewards often do not lead to a distribution that is amenable for sampling as above. The model distribution is thus instead used to propose samples, which in turn can yield low-quality (i.e. low-reward) samples especially due to the huge sequence space. This makes the exploration inefficient and even impractical.




Given the opposite behaviors of the algorithms in terms of exploration and computation efficiency, it is a natural idea to seek a middle ground between the two extremes in order to combine the advantages of both. 
Previous attempts have been made in this line from different perspectives.
We re-visit some of the popular approaches, and show that these apparently divergent algorithms can also be reformulated with the ERPO framework in Eqs.\eqref{eq:pg}-\eqref{eq:token-level-q}. 

\subsection{Reward-Augmented Maximum Likelihood (RAML)}\label{sec:raml}
RAML~\citep{norouzi2016reward} was originally proposed to incorporate task metric reward into the MLE training, and has shown superior performance. More formally, it introduces an exponentiated reward distribution $e(\bm{y}|\bm{y}^*)\propto \exp\{ R(\bm{y}|\bm{y}^*) \}$ where $R$ can be a task metric such as BLEU. RAML maximizes the following objective:
\begin{equation}
\small
\begin{split}
\mathcal{L}_{\text{RAML}}(\bm{\theta}) = \E_{\bm{y}\sim e(\bm{y}|\bm{y}^*)}\big[ \log p_\theta(\bm{y} ) \big].
\end{split}
\label{eq:raml}
\end{equation}
That is, unlike MLE that directly maximizes the data log-likelihood, RAML first perturbs the data proportionally to the reward distribution $e$, and maximizes the log-likelihood of the resulting samples. 

The RAML objective reduces to the vanilla MLE objective if we replace the task reward $R$ in $e(\bm{y}|\bm{y}^*)$ with the MLE $\delta$-reward from Eq.\eqref{eq:delta-reward}.
The relation between MLE and RAML still holds within our new formulation (Eqs.\ref{eq:pg}-\ref{eq:pg-em}). In particular, similar to how we recovered MLE from Eq.(\ref{eq:pg-em}), let $(\alpha\to 0, \beta=1)$\footnote{The exponentiated reward distribution $e$ can also include a temperature $\tau$~\citep{norouzi2016reward}. In this case, we set $\beta=\tau$.}, but set $R$ to the task metric reward, then the M-step of Eq.(\ref{eq:pg-em}) is precisely equivalent to maximizing the above RAML objective.

Having formulated RAML with the same framework, we can now compare it with other algorithms in the family.
In particular, similar to common policy optimization, the use of diffused task reward $R$ instead of $R_\delta$ permits a larger exploration space with valid reward signals. On the other hand, since $\alpha\to0$, by the form of $q$ in Eq.\eqref{eq:pg-em}, the model distribution $p_\theta(\bm{y})$ is not used for proposing samples. Thus the exploration space exposed for model training is only the regions surrounding the training examples as defined by the reward (Figure~\ref{fig:space}(b)). 
Besides, 
sampling from the reward-defined distribution, though tending to yield high-quality samples, can be difficult and require specialized techniques for efficiency~\citep[e.g.,][]{ma2017softmax}. 


\subsection{Softmax Policy Gradient (SPG)}\label{sec:spg}
SPG~\citep{ding2017cold} was developed in the perspective of adapting the vanilla policy gradient~\citep{sutton2000policy} to incorporate reward for proposing samples. SPG has the following objective:
\begin{equation}
\small
\begin{split}
\mathcal{L}_{SPG}(\bm{\theta}) = \log \E_{p_\theta}\left[ \exp R(\bm{y}|\bm{y}^*) \right],
\end{split}
\label{eq:spg}
\end{equation}
where $R$ is a task reward as above. 

The SPG algorithm can readily be fit into the ERPO framework. Specifically, taking gradient of Eq.(\ref{eq:spg}) w.r.t $\bm{\theta}$, we immediately get the same update rule as in Eq.(\ref{eq:pg-em}) with $(\alpha=1, \beta=0, R=\text{task reward})$. 

The only difference between the SPG and RAML configurations is that now $\alpha=1$.
SPG thus moves a step further than RAML by leveraging both the reward and the model distribution for full exploration (Figure~\ref{fig:space}(c)). Sufficient exploration at training would in principle boost the test-time performance. However, with the increased exploration and tendency of proposing lower-reward samples, additional  optimization and approximation techniques are usually needed~\citep{ding2017cold} to make the training practical.

\subsection{Data Noising}
Adding noise to training data is a widely adopted technique for regularizing models. Previous work~\citep{xie2017data} has proposed several data noising strategies in the sequence generation context. For example, a \emph{unigram noising}, with probability $\gamma$, replaces each token in data $\bm{y}^*$ with a sample from the unigram frequency distribution. The resulting noisy data is then used in MLE training.

Though previous literature has commonly seen such techniques as a data pre-processing step that differs from the above learning algorithms, we show the ERPO framework can also subsume data noising as a special instance. Specifically, starting from the ERPO reformulation of MLE which takes $(R=R_\delta, \alpha\to0, \beta=1)$ (section~\ref{sec:mle}), data noising can be formulated as using a locally relaxed variant of $R_\delta$. For example, assume $\bm{y}$ has the same length with $\bm{y}^*$ and let $\Delta_{\bm{y}, \bm{y}^*}$ be the set of tokens in $\bm{y}$ that differ from the corresponding tokens in $\bm{y}^*$, then a simple data noising strategy that randomly replaces a single token $y^*_t$ with another uniformly picked token is equivalent to using a reward $R'_\delta(\bm{y}|\bm{y}^*)$ that takes $1$ when $|\Delta_{\bm{y}, \bm{y}^*}|=1$ and $-\infty$ otherwise. Likewise, the above unigram noising~\citep{xie2017data} is equivalent to using a reward
\begin{equation}
\small
\begin{split}
R^{\text{unigram}}_\delta(\bm{y}|\bm{y}^*) = \left\{
  \begin{array}{ll}
  \log\left(\gamma^{|\Delta_{\bm{y}, \bm{y}^*}|}(1-\gamma)^{T-|\Delta_{\bm{y}, \bm{y}^*}|}\prod_{y_t\in\Delta_{\bm{y}, \bm{y}^*}} u(y_t) \right) & \text{if}\ \ T = T^* \\
  -\infty & \text{otherwise},
  \end{array}
  \right.
\end{split}
\label{eq:delta-reward-noise}
\end{equation}
where $u(\cdot)$ is the unigram frequency distribution.

With a relaxed reward, data noising expands exploration locally surrounding the training examples (Figure~\ref{fig:space}(b)). The effect is essentially the same as the RAML algorithm (section~\ref{sec:raml}), except that RAML expands exploration guided by the task reward.

\paragraph{Other Algorithms \& Discussions}
\citet{ranzato2015sequence} made an early attempt to mix the classic policy gradient algorithm~\citep{sutton2000policy} with MLE training. We show in the supplementary materials that the algorithm is closely related to our framework and can be recovered with moderate approximations. 
We have presented the framework in the context of sequence generation. The formulation can also be extended to other settings. For example, in game environments, $\bm{y}$ is a sequence of actions and states. 
The popular GAIL~\citep{ho2016generative} imitation learning approach uses an adversarially induced $R$ from data, and applies standard RL updates to train the policy. The policy update part can be formulated with our framework as standard policy gradient (with $\alpha=1, \beta=0$, and moderate approximation as above). The new algorithm described in the next section can also be applied to improve the vanilla GAIL, as shown in the experiments.

Previous work has also studied connections of relevant algorithms. For example, \citet{norouzi2016reward,koyamada2018alpha} formulate MLE and policy gradient as minimizing the opposite KL divergences between the model and data/reward distributions. \citet{misra2018policy} studied an update equation generalizing maximum marginal likelihood and policy gradient. Our framework differs in that we reformulate a different and more comprehensive set of algorithms for sequence prediction, and provide new insights in terms of exploration and its efficiency in a consistent view, which could not be derived from the previous work.
Section~\ref{sec:related} discusses more related work on sequence prediction learning.

\section{Interpolation Algorithm}\label{sec:alg}
The unified perspective enables new understandings of the existing algorithms, and can also facilitate new algorithms for further improved learning. Here we present an example algorithm that is naturally inspired from the framework.

As in Figure~\ref{fig:conn-dots}, each of the learning algorithms can be seen as a point in the $(R, \alpha, \beta)$ hyperparameter space. Generally, from left to right, the reward gets more diffused and $\alpha$ gets larger, which results in larger sequence space exposed for model training (Figure~\ref{fig:space}), while in turn making the training less efficient due to lower sample quality. 
We propose an \emph{interpolation} algorithm to exploit the natural idea of starting learning from the most restricted yet efficient problem configuration, and gradually expanding the exploration to decrease the training/test discrepancy. The easy-to-hard learning paradigm resembles the curriculum learning~\citep{bengio2009curriculum}.
As we have mapped the algorithms to points in the hyperparameter space, the interpolation becomes straightforward, which is reduced to simple \emph{annealing} of the hyperparameter values.

\begin{table}[t]
\centering
\small
\begin{tabular}{@{}r l@{}} 
\cmidrule[\heavyrulewidth]{1-2}
{\bf Model} & {\bf BLEU}\\ \cmidrule{1-2}
MIXER~\citep{ranzato2015sequence} & 21.83 \\ 
BSO~\citep{wiseman2016sequence} & 26.36 \\
Actor-critic~\citep{bahdanau2016actor} & 28.53 \\
Minimum-risk~\citep{edunov2017classical} & 32.84 \\
\cmidrule{1-2}
MLE & $31.99 \pm 0.17$ \\
RAML~\citep{norouzi2016reward} & $32.51 \pm 0.37$ \\
Self-critic~\citep{rennie2017self} & $32.23 \pm 0.15$  \\
Scheduled Sampling~\citep{bengio2015scheduled} & $32.13 \pm 0.14$ \\
{\bf Ours} & $\bm{33.35 \pm 0.08}$ \\
\cmidrule[\heavyrulewidth]{1-2}
\end{tabular}
\vspace{-10pt}
\captionof{table}{Machine translation results averaged over 5 runs.}
\label{tab:rst-mt}
\end{table}
\hfill
\begin{table}[t]
\centering
\small
\begin{tabular}{@{}r l l l@{}} 
\cmidrule[\heavyrulewidth]{1-4}
{\bf Method} & {\bf ROUGE-1} & {\bf ROUGE-2} & {\bf ROUGE-L}\\ \cmidrule{1-4}
MLE & $36.11 \pm 0.21$ & $16.39 \pm 0.16$ & $32.32 \pm 0.19$ \\
RAML~\citep{norouzi2016reward} & $36.30 \pm 0.04$ & $16.69 \pm 0.20$ & $32.49 \pm 0.17$ \\
Self-critic~\citep{rennie2017self} & $36.48 \pm 0.24$ & $16.84 \pm 0.26$ & $32.79 \pm 0.26$ \\
Scheduled Sampling~\citep{bengio2015scheduled} & $36.59 \pm 0.12$ & $16.79 \pm 0.22$ & $32.77 \pm 0.17$ \\
{\bf Ours} & \bm{$36.72 \pm 0.29$} & \bm{$16.99 \pm 0.17$} & \bm{$32.95 \pm 0.33$} \\
\cmidrule[\heavyrulewidth]{1-4}
\end{tabular}
\vspace{-10pt}
\captionof{table}{Text summarization results averaged over 5 runs.}
\label{tab:rst-summ}
\end{table}

Specifically, in the general update rule Eq.(\ref{eq:pg-em}), we would like to anneal from using $R_\delta$ to using task reward, and anneal from exploring by only $R$ to exploring by both $R$ and $p_\theta$. Let $R_{\comm}$ denote a task reward (e.g., BLEU). The interpolated reward can be written in the form $R = \lambda R_\comm + (1-\lambda) R_{\delta}$, for $\lambda\in[0,1]$. Plugging $R$ into $q$ in Eq.(\ref{eq:pg-em}) and re-organizing the scalar weights, we obtain the numerator of $q$ in the form: $c\cdot(\lambda_1 \log p_\theta + \lambda_2 R_\comm + \lambda_3 R_\delta)$, where $(\lambda_1, \lambda_2, \lambda_3)$ is defined as a distribution (i.e., $\lambda_1 + \lambda_2 + \lambda_3 = 1$), and, along with $c\in \mathbb{R}$, are determined by $(\alpha, \beta, \lambda)$. For example, $\lambda_1 = \alpha / (\alpha+1)$. We gradually increase $\lambda_1$ and $\lambda_2$ and decrease $\lambda_3$ as the training proceeds.

Further, noting that $R_\delta$ is a Delta function (Eq.\ref{eq:delta-reward}) which would make the above direct function interpolation problematic, we borrow the idea from the Bayesian \emph{spike-and-slab} factor selection method~\citep{ishwaran2005spike}. That is, we introduce a categorical random variable $z\in\{1,2,3\}$ that follows the distribution $(\lambda_1, \lambda_2, \lambda_3)$, and augment $q$ as $q(\bm{y}|z) \propto \exp\{c\cdot (\indicator(z=1)\log p_\theta + \indicator(z=2) R_\comm + \indicator(z=3) R_\delta) \}$. The M-step is then to maximize the objective with $z$ marginalized out: $\max_{\bm{\theta}} \E_{p(z)}\E_{q(\bm{y}|z)}\left[ \log p_\theta(\bm{y}) \right]$. The spike-and-slab adaption essentially transforms the product of experts in $q$ to a mixture,  which resembles the bang-bang rewarded SPG method~\citep{ding2017cold} where the name \emph{bang-bang} refers to a system that switches abruptly between extreme states (i.e., the $z$ values). Finally, similar to \citep{ding2017cold}, we adopt the token-level formulation (Eq.\ref{eq:token-level-q}) and associate each token with a separate variable $z$.

We provide the pseudo-code of the interpolation algorithm in the supplements. It is notable that \cite{ranzato2015sequence} also developed an annealing strategy that mixes MLE and policy gradient training. As discussed in the supplements, the algorithm can be seen as a special instance of the ERPO framework with moderate approximation. 

As discussed above, we can also apply the interpolation algorithm in game imitation learning, by plugging it into the GAIL~\citep{ho2016generative} framework to replace the standard RL routine for policy update. 
The annealing schedule is constrained due to the agent interaction with the environment. Specifically, to generate a trajectory (a sequence of actions and states), we sample the beginning part from data (demonstrations), followed by sampling from either the model or reward. Note that data sampling can happen only before model/reward sampling, because the latter will interact with the environment and result in states that do not necessarily match the data. Similar to sequence generation, we gradually anneal from data sampling to model/reward sampling, and hence increase the exploration until converging to standard RL. Our experiments validate that the easy-to-hard training is superior to the vanilla GAIL which directly applies the hard RL update from the beginning.

\section{Experiments}\label{sec:exp}
We evaluate the interpolation algorithm in the context of both text generation (including machine translation and summarization) and game agent learning. Experiments are run with 4 GTX 2080Ti GPUs and 32GB RAM. Code is included in the supplementary materials and will be cleaned and released upon acceptance.

\subsection{Machine Translation}
We use a state-of-the-art network architecture, namely, Transformer~\citep{vaswani2017attention} as the base model. Our transformer has 6 blocks. Adam optimization is used with an initial learning rate of 0.001 and the same schedule as in~\citep{vaswani2017attention}. Batch size is set to 1,792 tokens. At test time, we use beam search decoding with a beam width of 5 and length penalty 0.6.
We use the popular IWSLT2014~\citep{cettolo2014report} German-English dataset. After proper pre-processing as described in the supplementary materials, we obtain the final dataset with train/dev/test size of around 146K/7K/7K, respectively. The shared de-en vocabulary is of size 73,197 without BPE encoding.

Table~\ref{tab:rst-mt} shows the test-set BLEU scores of various methods. Besides MLE and RAML as described above, we also compare with Self-critic~\citep{rennie2017self}, an RL-based approach, as well as Scheduled Sampling (SS). 
As a reference, we also list the results from previous papers that proposed various learning algorithms (though with different model architectures).
From the table, we can see the various approaches such as RAML provide improved performance over the vanilla MLE, as more sufficient exploration is made at training time. Our interpolation algorithm performs best, with significant improvement over the MLE training by 1.36 BLEU points.
The results validate our approach that interpolates among the existing algorithms offers beneficial scheduled training. We note that there is other work exploring various network architectures for machine translation~\citep{shankar2019posterior,he2018layer}, which is orthogonal and complementary to the learning algorithms. It would be interesting to explore the effect of combining the approaches.



\subsection{Text Summarization}
We use an attentional sequence-to-sequence model~\citep{luong2015effective} where both the encoder and decoder are single-layer LSTM RNN. The dimensions of word embedding, RNN hidden state, and attention are all set to 256. 
We use Adam optimization for training, with an initial learning rate of 0.001 and batch size of 64. At test time, we use beam search decoding with a beam width of 5. 
Please see the supplementary materials for more configuration details.
%
We use the popular English Gigaword corpus~\citep{graff2003english} for text summarization, and pre-processed the data following~\citep{rush2015neural}. The resulting dataset consists of 200K/8K/2K source-target pairs in train/dev/test sets, respectively. 

Following previous work~\citep{ding2017cold}, we use the summation of the three ROUGE(-1, -2, -L) metrics as the reward in learning. Table~\ref{tab:rst-summ} show the results on the test set. The proposed interpolation algorithm achieves the best performance on all three metrics. The RAML algorithm, which performed well in machine translation, falls behind other algorithms in text summarization. In contrast, our method consistently provides the best results.

\subsection{Game Imitation Learning}
We apply the interpolation algorithm in GAIL~\citep{ho2016generative} as described in section~\ref{sec:alg}. 
Following~\citep{ho2016generative}, we simulate three environments with MuJoCo~\citep{todorov2012mujoco}. Expert demonstrations are generated by running PPO~\citep{schulman2017proximal} under the given true reward functions. We then run different imitation learning algorithms with varying numbers of demonstrations. Both the policy and the discriminator are two-layer networks with 128 units each and tanh activations in between.

\begin{figure*}[t]
    \centering
    \begin{subfigure}[b]{0.32\textwidth}
        \centering
        \includegraphics[width=\textwidth]{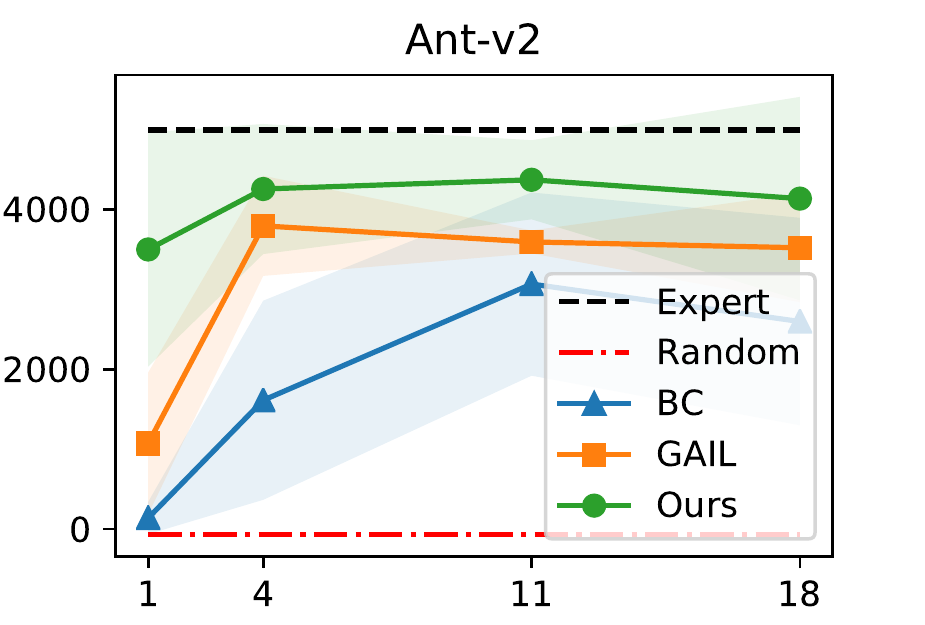}
    \end{subfigure}%
    \begin{subfigure}[b]{0.32\textwidth}
        \centering
        \includegraphics[width=\textwidth]{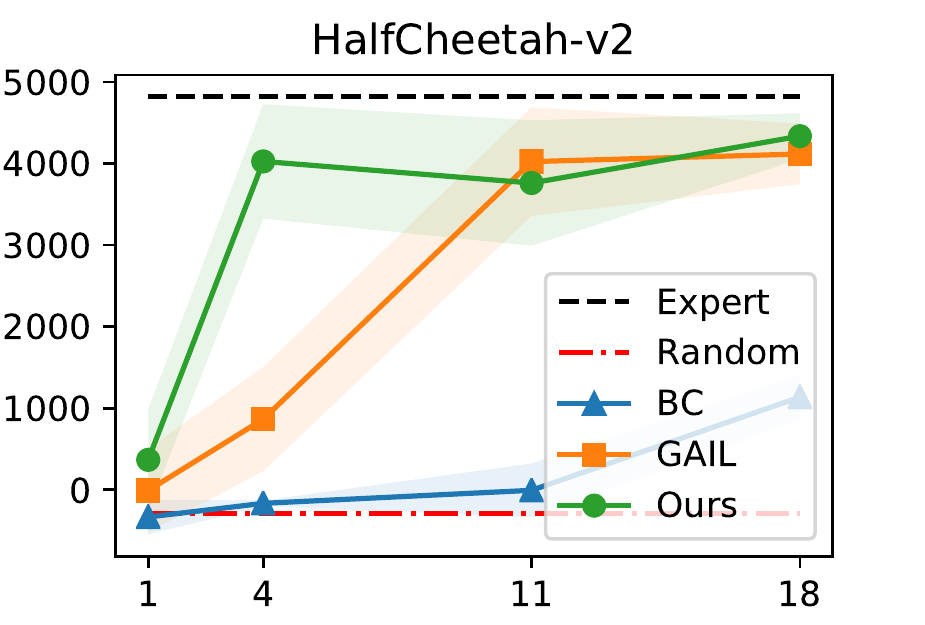}
    \end{subfigure}%
    \begin{subfigure}[b]{0.32\textwidth}
        \centering
        \includegraphics[width=\textwidth]{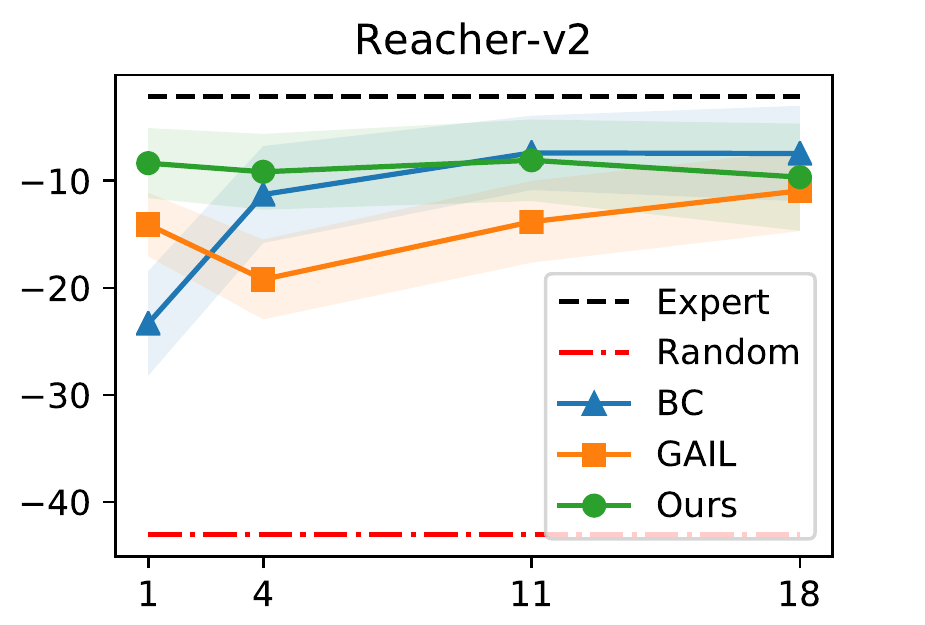}
    \end{subfigure}%
    \vspace{-5pt}
    \caption{Performance of learned policies. The x-axis is the number of expert demonstrations for training. The y-axis is the average returns. ``BC'' is Behavior Cloning. ``Random'' is a baseline taking a random action each time. Results are averaged over 50 runs.}
\label{fig:game}
\vspace{-3pt}
\end{figure*}

Figure~\ref{fig:game} shows the average returns by the agents. We can see that agents trained with the interpolation algorithm can generally improve over the vanilla GAIL, especially in the presence of small number (e.g., 1 or 4) of demonstrations. This shows that our approach that anneals from the MLE mode to RL mode can make better use of data examples, and steadily achieve better performance in the end.


\section{Conclusions}
We have presented a unified perspective of a variety of learning algorithms for sequence prediction problems. The framework is based on a generalized entropy regularized policy optimization formulation, and we show the distinct algorithms are mathematically equivalent to specifying certain hyperparameter configurations in the framework. The new consistent treatment provides systematic understanding and comparison across the algorithms, and inspires further improved learning. The proposed interpolation algorithm shows consistent improvement in machine translation, text summarization, and game imitation learning. 

\small
\bibliography{refs}
\bibliographystyle{abbrvnat}

\end{document}